**Neuro-Memory Fuzzy Inference System for Mimicking Human-like Car Following Behavior**

**Nazmul Haque**
Assistant Professor
Accident Research Institute (ARI)
Bangladesh University of Engineering and Technology (BUET)
Dhaka 1000, Bangladesh
Email: nhaque@ari.buet.ac.bd

**Md Asif Raihan***
Professor
Accident Research Institute (ARI)
Bangladesh University of Engineering and Technology (BUET)
Dhaka 1000, Bangladesh
Email: raihan@ari.buet.ac.bd

**Md. Hadiuzzaman**
Professor
Department of Civil Engineering
Bangladesh University of Engineering and Technology (BUET)
Dhaka 1000, Bangladesh
Email: mhadiuzzaman@ce.buet.ac.bd

***Corresponding Author**

## ABSTRACT

This study presents the Neuro-Memory Fuzzy Inference System (NeMeFIS), a hierarchical machine learning architecture that asymmetrically models acceleration and deceleration in car-following behavior by integrating five human memory types—procedural, working, episodic, semantic, and declarative. By linking external variables to memory functions via metaheuristics and validating them through factor and p-value analyses, NeMeFIS uncovers latent cognitive influences across Arterial, Collector, and Rural Highway corridors for different types of vehicles. Results from 54 different trained models emphasize cognitive thresholds shaped by driver perception limits and cognitive load. The trained NeMeFIS models outperform traditional statistical and conventional machine learning models in replicating realistic driving behavior, including comparisons with Linear Regression, ANFIS, and LSTM architectures. Fuzzy rule analysis reveals that declarative memory demands the highest rule, especially during deceleration, indicating complex braking decisions. Procedural memory drives acceleration, while semantic and declarative memory guide deceleration. Risk perception also emerges as a key factor, particularly on urban roads. Validated on both heterogeneous and homogeneous datasets, NeMeFIS offers a robust framework for modeling driver cognition. The findings support psychotherapeutic applications and the development of adaptive, human-like decision systems in Connected and Autonomous Vehicles (CAVs) to enhance traffic safety.

## INTRODUCTION

Car-Following (CF) models are essential for analyzing vehicle interactions and longitudinal dynamics using variables like relative speed, spacing, and velocity (Haque et al. 2024). They typically conceptualize the following vehicle as a reactor to stimuli from the leading vehicle, such as relative speed or desired spacing, with responses reflected in velocity or acceleration adjustments (Khodayari et al. 2010). These models support ITS applications, including Advanced Driver Assistance Systems (ADAS) and Automated Driving Systems (ADS), facilitating advancements in automated and connected vehicles by modeling driver behavior and transferring human skills to intelligent systems (Khodayari et al. 2011). As ITS technologies evolve, understanding vehicle interaction under these systems is vital (Ma and Andreasson 2007). While CF models often define vehicles as "Followers" and "Leaders," real-world traffic involves complex interactions with multiple vehicles, requiring drivers to make continuous decisions for safety and collision avoidance, demanding significant cognitive and psychological engagement (Ranney 1999).

Human memory plays a pivotal role in driving decisions, influencing actions based on past experiences, learned knowledge, and intuitive reasoning (Dougherty et al. 1999; Schooler and Hertwig 2005; Shadlen and Shohamy 2016). However, incorporating human memory into predictive models has been challenging, as it requires capturing intricate mental representations essential for simulating human-like driving behaviors (Zhao et al. 2022). Early CF models, such as Wiedemann (1974) and Fritzsche (1994), introduced psycho-physical thresholds to address perceptual limits, but these models failed to comprehensively account for the psychological processes underlying driver behavior. Boer (1999) criticized these models for their simplification of human behavior and assumption of optimal driver performance. Later advancements integrated factors such as visual angles (Jin et al. 2011) , driving risk (Hamdar et al. 2008 ; Hamdar et al. 2015 ; Talebpour et al. 2011) , and distractions or errors (Lee et al. 2008 ; Treiber et al. 2006). Machine learning techniques, including Artificial Neural Networks (ANN) (Jia et al. 2003) and Adaptive Neuro-Fuzzy Inference Systems (ANFIS) (Chakroborty and Kikuchi 2003), further enriched the field. More recently, deep learning approaches like Recurrent Neural Networks (RNN) (Zhou et al. 2017), Long Short-Term Memory (LSTM) networks (Huang et al. 2018), Gated Recurrent Units (GRU) (Tang et al. 2020), and Deep Q-learning (Tang et al. 2022) have demonstrated potential in enhancing CF models. Chen et al. (2020) introduced Long and Short-Term Driving (LSTD) characteristics to explicitly incorporate human factors.

As Connected Autonomous Vehicles (CAVs) become increasingly prevalent, there is a growing imperative to design systems that mimic human-like decision-making behaviors. Current CAVs primarily rely on rule-based algorithms and machine learning models to perform tasks such as acceleration, braking, and lane changes. While these systems excel in efficiency, safety, and compliance with traffic regulations, they often overlook the complex, adaptive nature of human decision-making, particularly in unpredictable real-world scenarios (Shalev-Shwartz et al. 2017). Human drivers account for various factors—empathy, social context, and situational awareness—when making decisions such as when to yield, how to react to emergencies, and how to navigate intricate traffic conditions (Gogoll and Müller 2017). CAVs, by contrast, lack the ability to replicate these detailed behaviors, which can lead to decisions that, while technically safe, may fail to consider the emotional and social impacts on road users (Maurer et al. 2016). For instance, rigid adherence to safety distances or traffic rules may inadvertently increase stress among human drivers or pedestrians, highlighting the need for CAVs to incorporate human-like behaviors that foster safer, more compassionate interactions. Furthermore, aligning CAV behavior with societal values and ethical considerations is vital for gaining public trust in autonomous driving technologies (Gogoll and Müller 2017; Maurer et al. 2016).

To address these challenges, this study undertakes the development of a Neuro-Memory Fuzzy Inference System (NeMeFIS), a hierarchical machine learning architecture based model that asymmetrically simulates acceleration and deceleration behaviors by integrating five distinct types of human memory—procedural, working, episodic, semantic, and declarative. The model links memory functions to external driving variables using metaheuristic optimization and validates them through factor analysis and p-value assessments. NeMeFIS is trained, tested, and evaluated using UAV-captured naturalistic trajectory data across heterogeneous road environments, including urban arterials, collectors, and rural highways. The architecture allows disaggregated modeling across vehicle types and roadway contexts, enabling the identification of cognitive thresholds and memory-specific rule bases that govern magnitude decisions. Furthermore, numerical simulations confirm the stability and realism of the generated trajectories, establishing NeMeFIS as a robust cognitive framework for mimicking human-like driving.

The remainder of this paper is organized as follows: Section II reviews existing literature relevant to car-following behavior and cognitive modeling. Section III introduces the NeMeFIS architecture, detailing its hierarchical structure and the integration of memory types. Section IV describes the data acquisition process, preprocessing techniques, and analytical procedures. Section V presents the training, validation, and testing of the NeMeFIS model, alongside numerical simulations and performance evaluation. Section VI concludes the study with key findings, practical implications, and potential applications.

## RELEVANT WORKS

### Psychology-uninformed Car-Following Models

Traditional car-following models, such as the Gazis–Herman–Rothery (GHR) model (Chandler et al. 1958) , rely on stimulus-response frameworks that predict acceleration based on speed, relative speed, and spacing—largely omitting psychological dimensions. Enhancements like the Intelligent Driver Model (Treiber et al. 2000) , Gipps' model (Gipps 1981) , and Bando's model (Bando et al. 1995) focused on stability and numerical simulation but still lacked behavioral depth. A notable exception is Hamdar et al. (2008) , who incorporated Prospect Theory to model risk-taking under ambiguity.

To overcome these limitations, Machine Learning (ML) methods have gained traction for modeling car-following behavior by capturing nonlinearities and latent psychological factors. Jia et al. (2003) applied a backpropagation neural network using speed and spacing to classify driver risk levels, although validation was limited. Panwai and Dia [33] used a Radial Basis Function Network (RBFN) and fuzzy ARTMAP, mapping perception to action in stop-and-go traffic, but did not model reaction times explicitly. Huang et al. (2018) introduced an LSTM neural network to encode driving memory, improving traffic flow realism, albeit with high data demands. Aghabayk et al. (2013) adopted the LOLIMOT neuro-fuzzy system to reflect perceptual imperfections, revealing the challenge of fuzzy rule complexity and data dependency. More recently, these developments emphasize the shift toward psychologically grounded, data-driven modeling to bridge the gap between realism and interpretability in driving behavior models.

### Psychology-informed Car-Following Models

Lee (1966) extended the linear GHR model by incorporating a statistical memory function to capture a driver's reaction to relative speed over time, though it aggregated behavior using basic statistical parameters. Wiedemann (1974) later introduced perceptual thresholds to define behavioral regimes (free-flow, following, emergency) using spacing and speed, marking a psycho-physical advancement. However, his model lacked cognitive components and faced issues like perceptual switching without stimuli.

ML models have since emerged to implicitly encode psychological factors. Fuzzy logic-based approaches by Chakroborty and Kikuchi (2003), Kikuchi and Chakroborty (1992), and Khodayari et al. (2011) mapped external stimuli to driver responses using fuzzy inference systems. The Fadhloun-Rakha model advanced this by integrating randomness in perception and vehicle dynamics (Fadhloun and Rakha 2020). Huang et al. (2018) used LSTM networks to model memory-informed behavior based on past patterns. Hamdar et al. (2008) and Li et al. (2021) incorporated Prospect Theory and fuzzy logic to model decision-making under uncertainty.

Several models have explicitly addressed cognitive mechanisms. Talebpour et al. (2011) introduced a regime-switching model grounded in psychological decision theories. Tang et al. (2014) extended FVD models to incorporate driving styles and their impact on traffic flow. Jia et al. (2008) proposed a cognitive behavior model structured around perception, memory, and decisions, though lacking empirical validation. van Winsum (1999) modeled adaptive headways under fatigue, while Furutani (1976) introduced "following defense" behavior via catastrophe theory. Yi-Rong et al. (2015) integrated anticipation and response delays.

Despite progress, limitations remain in interpretability, data demands, and realistic psychological integration, underscoring future research needs.

## METHODOLOGY

### Hierarchical Fuzzy Systems

Fuzzy logic systems offer transparency in interpretation and analysis, making them suitable for complex applications. However, conventional fuzzy systems, while being universal approximators—suffer from the "curse of dimensionality," where increasing input parameters exponentially increases rule count, parameters, and data requirements. This often leads to overfitting and loss of system generalizability and transparency. To address this, Raju, Zhou, and Kisner introduced the concept of hierarchical fuzzy systems in the early 1990s (Wang et al. 2006). These systems decompose high-dimensional problems into interconnected, low-dimensional fuzzy logic units organized hierarchically. Outputs from one level serve as inputs for subsequent levels, enabling scalability and reducing complexity. He described such systems using tree structures with multiple levels and units. Mathematically, as described by Sun and Huo (2016), the first unit processes two real inputs, while subsequent units incrementally integrate previous outputs with new inputs, enhancing model interpretability and performance. This process advances till all the real inputs have been used.

Assuming there are $n$-input parameters $\{X_1, X_2, X_3, \dots X_n\}$ and $\{\widehat{X_1}, \widehat{X_2}, \widehat{X_3}, \dots \widehat{X_n}\}$ are the fuzzy variables extracted from input variables. 'm' presents total fuzzy rules, membership functions for inputs and outputs are presented by $M_{U_i^{j(x_i)}}$ and $M_{O_i^{j(x_i)}}$. With singleton fuzzifier and centroid de-fuzzifier, the hierarchical system can be represented below by equation Eq. (1) and (2).

$$IF\ (\widehat{X_1} = U_1^j)\ AND\ (\widehat{X_2} = U_2^j)\ THEN\ (Y_1 = O_1^j) \tag{1}$$

$$IF\ (\widehat{X_{i+1}} = U_{i+1}^j)\ AND\ (\widehat{Y_{i-1}} = O_{i-1}^j)\ THEN\ (Y_i = O_i^j) \tag{2}$$

where, '$j = \{1, 2, 3, \dots, m\}$; $U_1^j$ and $U_2^j$; $O_i^{j)}$ represents the fuzzy sets of linguistic labels for input and outputs respectively; $i = \{1, 2, 3, \dots, n-1\}$. $\widehat{X_1}$ and $\widehat{X_2}$ present the real-input variable and output from the previous fuzzy logic unit respectively. The Output at different level can be recursively calculated using the Eq. (3) and (4).

$$Y_1 = \frac{\sum_{j=1}^{m1} Q_{1,j} M_{U_1^j}(X_i)\, M_{U_2^j}(X_i)}{\sum_{j=1}^{m1} M_{U_1^j}(X_i)\, M_{U_2^j}(X_i)} \tag{3}$$

$$Y_i = \frac{\sum_{j=1}^{m1} Q_{i,j} M_{O_{i-1}^j}(Y_{i-1})\, M_{U_{i+1}^j}(X_{i+1})}{\sum_{j=1}^{m1} M_{O_{i-1}^j}(Y_{i-1})\, M_{U_{i+1}^j}(X_{i+1})} \tag{4}$$

$M$ ($M_{U_{i,j}}$ or $M_{O_{i,j}}$) can adopt any membership function. If bell shaped membership function is employed, $m_{A_{i,j}}$ is given by Eq. (5),

$$M_{U_{i,j}}(X_i) = \frac{1}{1 + \left[\left(\frac{X_i - c_{i,j}}{a_{i,j}}\right)^2\right] b_{i,j}}, i = 1,2, \dots \dots n, j = 1,2, \dots \dots, R \tag{5}$$

or the Gaussian membership function by Eq. (6),

$$M_{U_{i,j}}(X_i) = exp\left[-\left(\frac{X_i - c_{i,j}}{a_{i,j}}\right)^2\right], i = 1,2, \dots \dots n, j = 1,2, \dots \dots, R \tag{6}$$

where $a_{i,j}, b_{i,j}, c_{i,j}$ are the parameters of the membership functions corresponding to input $X_i$. The parameters in this layer are usually referred to as premise parameters. $Q_{i,j}$ is the normalized firing strength.

**The Magnitude Model: Neuro-Memory Fuzzy Inference System (NeMeFIS)**

For easiness of application the study considers 3 broader types of memory 1) Sensory 2) Declarative Memory and 3) Non-Declarative Memory. The Declarative Memory are categorized into three types 1) Working memory 2) Episodic Memory and 3) Semantic memory; as per Matthews (2015) and Wolk and Budson (2010). The NeMeFIS (Neuro-Memory Fuzzy Inference System) architecture shown in **Figure 1**, integrates these human memory types by including Procedural Memory (P), Working Memory (W), Episodic Memory (E), Semantic Memory (S), Declarative Memory (D), and Combined Memory (C). The system is designed to mimic human-like decision-making processes by mapping the functionality of each memory type into fuzzy inference systems (FIS) that process inputs and provide outputs based on specific memory roles.

The architecture consists of multiple subsystems (**Figure 1**), each representing a memory type, such as P-FIS for Procedural Memory, W-FIS for Working Memory, and so on. Inputs to the system are represented as feature vectors ($X \in \{\mathbf{X}_P, \mathbf{X}_W, \mathbf{X}_E, \mathbf{X}_S, \mathbf{X}_D, \mathbf{X}_C\}$), each containing multiple variables related to their corresponding memory types. The whole architechture is moduled into several Fuzzy Inference System (FIS). The P-FIS, W-FIS, E-FIS and S-FIS are representative of Procedural, Working, Episodic and Semantic memory. The W-FIS, E-FIS and S-FIS connects to D-FIS for Declarative memory. And the P-FIS and the D-FIS connects to C-FIS, the combiner of the memories. Each FIS subsystem computes a specific output ($Y \in \{\mathbf{y}_P, \mathbf{y}_w, \mathbf{y}_E, \mathbf{y}_S, \mathbf{y}_D, \mathbf{y}_C\}$) based on its memory function. Procedural Memory (P-FIS) handles habitual tasks and routine actions, while Working Memory (W-FIS) processes information in real-time. Episodic Memory (E-FIS) captures contextual and time-specific information, and Semantic Memory (S-FIS) processes factual and generalized knowledge. Declarative Memory (D-FIS) integrates knowledge from Episodic and Semantic Memories, providing a consolidated representation of explicit knowledge. Combined Memory (C-FIS) further integrates outputs from Declarative Memory and other subsystems, producing a final output that encapsulates the decision-making process across all memory types.

In NeMeFIS, driving memory is encoded during training through the creation and tuning of fuzzy rules and their associated membership functions, each corresponding to one of the five memory types (procedural, working, episodic, semantic, and declarative). As the model learns from vehicle-following data, it clusters input–output patterns—such as spacing, speed, and angular relationships—into distinct rule sets whose parameters (e.g., rule weights and membership-function shapes) capture habitual responses, short-term adaptations, event-specific recollections, generalized knowledge, and integrated cognitive streams. These rule parameters and membership-function boundaries collectively form the "memory" surface: during inference, each new driving situation activates the relevant rules in proportion to how closely current inputs match the stored prototypes, thereby reproducing learned decision patterns.

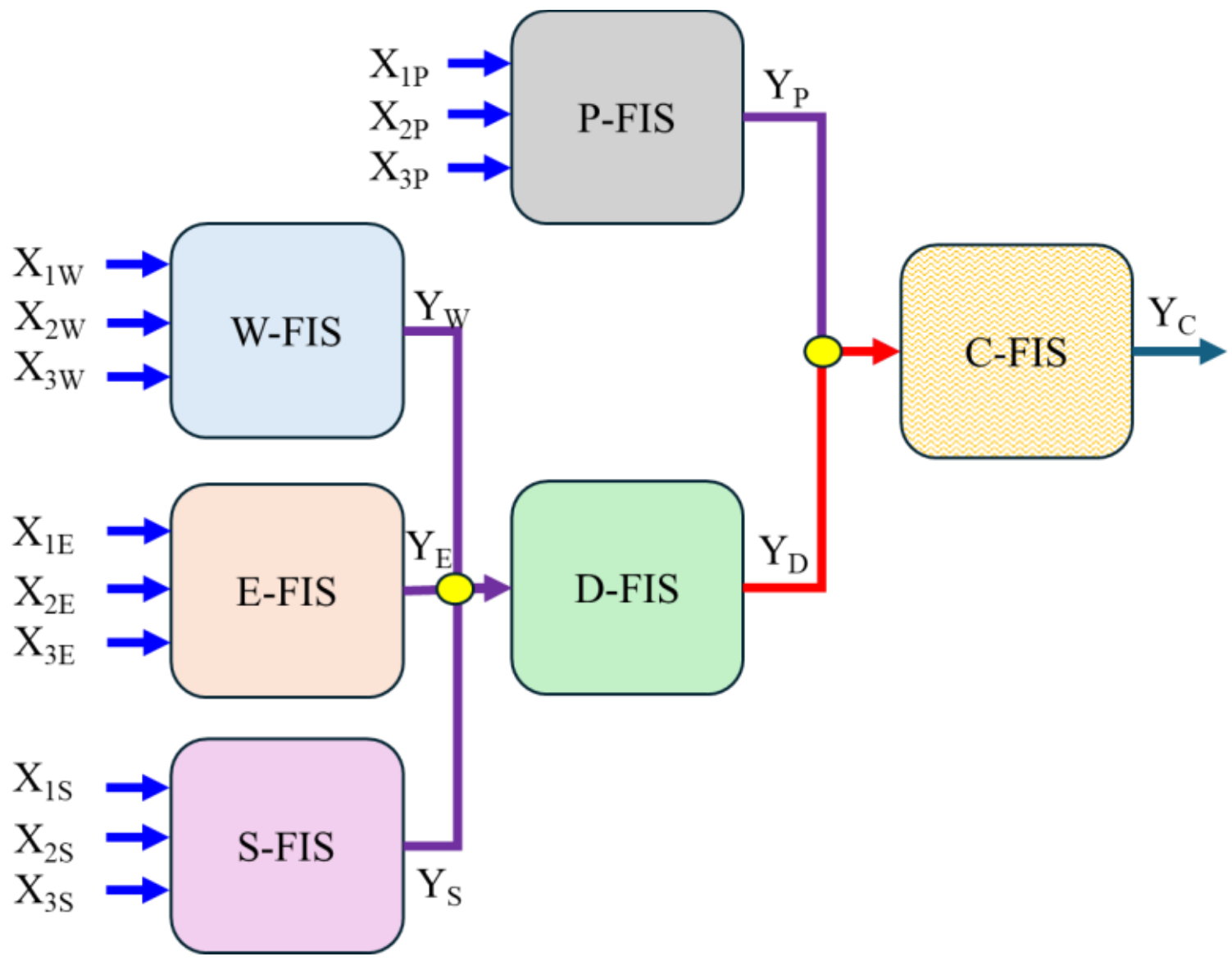

**Figure 1** Neuro Memory Fuzzy Inference System (NeMeFIS) Architecture

## DATA COLLECTION, PROCESSING AND ANALYSIS

### Study locations

This study employs a systematic data collection strategy, focusing on three road hierarchies of Dhaka, Bangladesh. The Arterial Road (AR) also features an 8-lane (3.65 m lane width) configuration and primarily accommodates private passenger cars (*Lat: 23.833542490523236, Long: 90.41936670104396*). The traffic flow rate during the experiment in this corridor was 4522 pcu/hr towards northbound and 4460 pcu/hr towards southbound. The average speed of the traffic stream (AR) was 33.41 kmph. The Collector Road (CR) comprises an 8-lane thoroughfare with four lanes (3.1 m lane width) in each direction, predominantly serving non-motorized transportation, primarily rickshaws (*Lat: 23.73904181123112, Long: 90.40997296817643*). The traffic flow rate during the experiment in this corridor was 4542 pcu/hr towards northbound and 4214 pcu/hr towards southbound. The average speed of the traffic stream (CR) was 20.77 kmph. The Rural Highway (RH) has a four-lane configuration with two lanes (3.65m lane width) in each direction, where inter-district buses are the primary mode of transportation (*Lat: 23.69844939827767, Long: 90.51082733705677*). The traffic flow rate during the experiment in this corridor was 2748 pcu/hr towards northbound and 2453 pcu/hr towards southbound. The average speed of this traffic stream (RH) was found to be 45.4 kmph.

### Experimental Set-up and Flight Plan

Video Data has been collected at 30 FPS using a DJI Mini 2 drone weighing 249 grams, equipped with a 4K HDR video recording capability and a 12MP camera sensor. The video resolution has been set to Full HD (1920 x 1080) at 32 Mbps bitrate to ensure optimal video quality for detection and classification while minimizing energy consumption. The data collection process occurred during morning peak hours from 9:00 AM to 10:00 AM on three random weekdays (*AR: 8th January 2019; CR: 30th January 2023; RH: 18th February 2023*).

Ensuring clear visibility and avoiding unnecessary noise in the data, days with extreme fog have been avoided in the data collection schedule. During the experiments, the drone has been flown at an altitude ranging between 23 to 27 meters (similar to the NGSIM (2007))and the viewing angle with the horizontal plane has been maintained at approximately 17 to 18 degrees. The flight has been initiated from high points at least 20 meters away from the road to minimize the impact on traffic and facilitate quick battery changes. Necessary field measurements have been taken for camera calibration and skew axis correction.

Each detected road user has a certain size represented by its bounding box information (vehicle type, centroid, width, and height) and is extracted using the framework DEEGITS proposed by Islam, Haque, and Hadiuzzaman (2024) for heterogeneous traffic stream. For simplicity, the centroid point has been taken as a simpler representation of the detected vehicle. The measurement of each road user's position in space at tracked time frame constitutes its trajectory. represents the trajectory of -th road users, which is a zipped collection of all points in the desired time interval. The microscopic measurements among the subject vehicle and the lead vehicle within the -vehicle interaction system are described below:

$$V_i(t) = d\big(T_i(t), T_i(t-\Delta t)\big)/\Delta t \quad (7)$$

$$a_i(t) = \big(V_i(t-\Delta t) - V_i(t)\big)/\Delta t \quad (8)$$

$$h_{ij}(x, y) = T_i(x, y) - T_j(x, y), \forall i: i < j \quad (9)$$

$$\Delta x_{ij}(t) = d\left(T_i(t), T_j(t)\right), \forall i: i < j \quad (10)$$

$$\Delta V_{ij}(t) = V_i(t) - V_j(t), \forall i: i < j \quad (11)$$

$$\theta_{ij}(t) = \tan^{-1}\frac{\left|\mathbf{Q}(T_i(t))\times\mathbf{R}\left(T_{ij}(t)\right)\right|}{\mathbf{Q}(T_i(t)).\mathbf{R}\left(T_{ij}(t)\right)}, \theta_{ij} \in [0, 2\pi] \quad (12)$$

$$\Delta y_{ij}(t) = \Delta x_{ij}(t)\cos\theta_{ij}(t) \quad (13)$$

$$TTC = \Delta x_{ij}(t)/\Delta V_{ij}(t) \quad (14)$$

$$RP = \frac{(NTTC - \max(NTTC))}{(\max(NTTC) - \min(NTTC))}, NTTC = P_{10}(\omega_k) \quad (15)$$

Equations (7) to (15) are for measuring speed, acceleration/deceleration, time headway, relative spacing, relative speed, lateral gap, time to collision and risk perception (Haque et al. 2025) respectively. $V_i$ and $V_j$ are the speed of the corresponding vehicles $i$ and $j$, $\Delta t$ is time interval in between two consecutive positions of the vehicles, $\mathbf{Q}$ and $\mathbf{R}$ are the direction vectors. $d(.)$ is the Euclidean distance function.

**Data Analysis**

**Table 1** presents the correlation between nine external variables and the response variable—Relative Speed—across three corridor types (AR, CR, RH) for both acceleration (A) and deceleration (D) phases. The results reveal a strong asymmetry in how variables influence acceleration versus deceleration. Speed shows the most dominant pattern, exhibiting strong positive correlations during acceleration (e.g., 0.103 in AR, 0.134 in CR) and near-perfect negative correlations during deceleration (–0.987 in AR, –0.988 in CR, –0.967 in RH), indicating that higher speeds sharply intensify braking responses across all corridor types. Relative speed (Var 8) also plays a major role but in the opposite direction: it shows strong positive correlations during deceleration (0.826 in AR, 0.758 in CR, 0.764 in RH), confirming that larger closing speeds trigger stronger deceleration demands. In contrast, spacing-related variables (Vars 1, 2, and 5) consistently show positive correlations during acceleration and moderate negative correlations during deceleration, highlighting that larger headways promote acceleration whereas reduced spacing induces braking. Angular variables (Vars 3 and 4) display weak correlations in most cases, suggesting limited direct influence on relative speed adjustments. Risk perception (Var 9) remains weakly correlated across all scenarios, indicating a more nuanced or indirect effect on moment-to-moment speed regulation.

**TABLE 1 Correlation of the External Variables with the Response Variable (i.e. Relative Speed)**

| | AR | | CR | | RH | |
|---|---|---|---|---|---|---|
| **Var** | **A** | **D** | **A** | **D** | **A** | **D** |
| 1 | **0.215** | **-0.321** | **0.282** | **-0.230** | **0.187** | **-0.228** |
| 2 | **-0.022** | **-0.299** | **0.061** | **-0.219** | **0.014** | **-0.279** |
| 3 | **0.007** | **0.000** | **0.002** | **-0.005** | **-0.002** | **0.014** |
| 4 | **0.065** | **-0.113** | **0.027** | **-0.058** | **-0.020** | **-0.074** |
| 5 | **0.262** | **-0.332** | **0.129** | **-0.235** | **0.130** | **-0.212** |
| 6 | **-0.002** | **-0.326** | **0.010** | **-0.213** | **0.017** | **-0.176** |
| 7 | **0.103** | **-0.987** | **0.134** | **-0.988** | **0.063** | **-0.967** |
| 8 | **-0.014** | **0.826** | **0.033** | **0.758** | **0.059** | **0.764** |
| 9 | **0.013** | **0.024** | **-0.025** | **0.018** | **0.053** | **-0.016** |

***Corridor Types:*** *AR = Arterial Road, CR = Collector Road, RH = Rural Highway; A=Acceleration, D=Deceleration, 1*= Spacing S-L ***, 2*** = Spacing S-NL ***, 3***= Angle S-L ***, 4*** = Angle L-NL ***, 5***= Spacing Change S-L, ***6*** = Spacing Change S-NL ***, 7***= Speed S ***8*** = Relative Speed S-NL ***, 9***= Risk Perception; *S-L = Subject-Leader, S-NL = Subject-Next Leader, L-NL = Leader-Next Leader, S = Subject Vehicle*

Traffic hysteresis phenomena is widely used to identify the trajectory pairs in the following behavior [51].

Understanding hysteresis allows researchers to properly categorize trajectory data for behavioral modeling, thereby improving the accuracy and reliability of analyses of vehicular interactions. Accordingly, hysteresis patterns were examined for all trajectory pairs to ensure that they conform to true following behavior.

**Integrating External Variables into Human Memory-Based Architecture**

The grouping of variables into Procedural, Working, Episodic, and Semantic Memory reflects their theoretical roles in driving behavior and is statistically supported by factor analysis. Procedural Memory variables such as Spacing S-L and S-NL show strong loadings in the AR corridor (0.565 and 0.948), confirming their habitual monitoring function. Episodic Memory variables—Spacing Change S-L and S-NL—exhibit high loadings in the RH corridor (0.697 and 0.736), indicating their relevance for recalling recent spacing variations. Semantic Memory is validated by the very strong loading of Speed S (0.992) in the CR corridor, highlighting its role in knowledge-driven, anticipatory decisions. Further justification is presented in Table II and the subsequent discussion.

**TABLE 2 Rationale Behind the Integrating Variables with Corresponding Memory Type**

| Memory Type | Variables | Justification |
|---|---|---|
| **Procedural Memory ($\mathbf{X}_P$)** | 1-Spacing S-L ($\Delta x_{i,i-1}$),<br>2-Spacing S-NL ($\Delta x_{i,i-2}$) | Habitual and intuitive maintenance of safe spacing, strongly loaded on factors with low uniqueness values. |
| **Working Memory ($\mathbf{X}_W$)** | 3-Angle S-L ($\theta_{i,i-1}$),<br>4-Angle L-NL ($\theta_{i,i-2}$) | Real-time spatial awareness and decision-making, supported by moderate factor loadings and specificity to active tasks. |
| **Episodic Memory( $\mathbf{X}_E$)** | 5-Spacing Change S-L ($(\Delta x_{i,i-1})/\Delta t$),<br>6-Spacing Change S-NL ($(\Delta x_{i,i-2})/\Delta t$) | Interpretation of dynamic spacing changes, associated with event-based memory of recent traffic changes. |
| **Semantic Memory($\mathbf{X}_S$)** | 7-Speed S ($V_i$),<br>8-Relative Speed S-NL ($\Delta V_{i,i-2}$),<br>9-Risk Perception ($R_{i,i-1}$) | General knowledge of speed and risk, reflecting abstract cognitive processes with high factor loadings. |

**Table 3** summarizes the statistical significance of each external variable across Arterial (AR), Collector (CR), and Rural Highway (RH) corridors using p-value distributions. Variables such as spacing (Var 1), spacing change (Var 5–6), and speed (Var 7) show a high count of p-values < 0.05, indicating strong and consistent predictive influence on relative speed. Angle-related variables (Var 3–4) generally have higher average p-values and fewer significant cases, suggesting weaker or more context-specific effects. Speed (Var 7) is universally significant in CR and RH (14 out of 14 cases), confirming its dominant role in car-following behavior.

**TABLE 3 P-Value Analysis of the External Variables**

| CT | Var | Min | Max | Avg. | Std. | 2.5% | 97.5% | #P-Value < 0.05 |
|---|---|---|---|---|---|---|---|---|
| AR | 1 | 0.00 | 0.86 | 0.19 | 0.33 | 0.00 | 0.86 | 10 |
| | 2 | 0.00 | 0.85 | 0.22 | 0.27 | 0.00 | 0.78 | 7 |
| | 3 | 0.09 | 0.93 | 0.49 | 0.26 | 0.11 | 0.89 | 0 |
| | 4 | 0.00 | 0.79 | 0.17 | 0.26 | 0.00 | 0.72 | 9 |
| | 5 | 0.00 | 0.79 | 0.12 | 0.24 | 0.00 | 0.73 | 10 |
| | 6 | 0.00 | 0.71 | 0.16 | 0.22 | 0.00 | 0.67 | 7 |
| | 7 | 0.00 | 0.88 | 0.06 | 0.23 | 0.00 | 0.60 | 13 |
| | 8 | 0.00 | 0.96 | 0.40 | 0.39 | 0.00 | 0.93 | 5 |
| | 9 | 0.00 | 0.88 | 0.33 | 0.27 | 0.00 | 0.84 | 3 |
| CR | 1 | 0.00 | 0.66 | 0.08 | 0.18 | 0.00 | 0.56 | 12 |
| | 2 | 0.00 | 0.80 | 0.16 | 0.26 | 0.00 | 0.72 | 9 |
| | 3 | 0.00 | 0.89 | 0.31 | 0.26 | 0.00 | 0.80 | 4 |
| | 4 | 0.00 | 0.93 | 0.20 | 0.30 | 0.00 | 0.88 | 8 |
| | 5 | 0.00 | 0.72 | 0.10 | 0.24 | 0.00 | 0.70 | 12 |
| | 6 | 0.00 | 0.55 | 0.10 | 0.18 | 0.00 | 0.54 | 10 |

| CT | Var | Min | Max | Avg. | Std. | 2.5% | 97.5% | #P-Value < 0.05 |
|---|---|---|---|---|---|---|---|---|
| | 7 | 0.00 | 0.00 | 0.00 | 0.00 | 0.00 | 0.00 | 14 |
| | 8 | 0.00 | 0.86 | 0.13 | 0.24 | 0.00 | 0.73 | 10 |
| | 9 | 0.00 | 0.86 | 0.09 | 0.22 | 0.00 | 0.65 | 11 |
| RH | 1 | 0.00 | 0.85 | 0.08 | 0.21 | 0.00 | 0.65 | 12 |
| | 2 | 0.00 | 0.75 | 0.14 | 0.23 | 0.00 | 0.68 | 10 |
| | 3 | 0.02 | 0.95 | 0.35 | 0.33 | 0.02 | 0.93 | 3 |
| | 4 | 0.00 | 0.96 | 0.41 | 0.39 | 0.00 | 0.95 | 6 |
| | 5 | 0.00 | 0.96 | 0.20 | 0.32 | 0.00 | 0.88 | 10 |
| | 6 | 0.00 | 0.93 | 0.18 | 0.31 | 0.00 | 0.93 | 9 |
| | 7 | 0.00 | 0.38 | 0.05 | 0.11 | 0.00 | 0.35 | 14 |
| | 8 | 0.00 | 0.90 | 0.18 | 0.25 | 0.00 | 0.76 | 8 |
| | 9 | 0.00 | 1.00 | 0.27 | 0.35 | 0.00 | 0.97 | 8 |

**See Table I for detailed notations*

## MODEL DEVELOPMENT AND ASSESSMENTS

The data is split into a training set (70%), a validation set (15%), and a testing set (15%). This randomized division assigns indices to each subset, ensuring no overlap among them. The training set is used to train the architecture, the validation set fine-tunes the model's hyperparameters to enhance its generalization capabilities, and the testing set evaluates the final model's performance on unseen data. For all vehicle types, the system employs five input membership functions, eight output membership functions, 30 combined membership functions, and 36 final membership functions. However, when the focus shifts to individual vehicle types, the parameters adjust to account for heterogeneity: three input MFs, six output MFs, 24 combined MFs, and the same 36 final MFs.

The use of Sugeno-type fuzzy inference systems (SugFIS) across all memory. Gaussian MFs dominate the design for their smoothness and tunability. Each FIS defines its input ranges based on observed data variability, ensuring adaptability to dynamic traffic scenarios. During the training phase, the hierarchical FIS structure, which is designed to represent different types of human memory (e.g., Procedural, Working, Episodic, Semantic, and Declarative), is optimized using Particle Swarm Optimization (Kennedy and Eberhart 1995). The validation phase further refines the trained FIS using the validation dataset. This step adopts a tuning-based approach, employing the Pattern Search optimization (Torczon 1997) method to fine-tune the hyperparameters and ensure the model generalizes well. Finally, the optimized architecture is tested on the testing dataset to evaluate its performance on new, unseen data. Predicted outputs are generated using the test inputs, and the RMSE is calculated to quantify the model's generalization capability. The training-tuning process converges in all three corridors, i.e., AR, CR and RH. Results showed that trained, tested and validated models can adopt the relative speed fluctuations with the parameter sets considered in AR corridor for acceleration and deceleration decisions.

**Table 4** presents the Root Mean Square Error (RMSE) for the model's predictions (i.e. normalized Relative Speed (m/s)) across various vehicle types and traffic corridors, categorized by training, testing, and validation. It took 385 hours to train, test, and validate the 54 models on Core i7-10700 CPUs with 24 GB RAM in parallel processing. The results demonstrate that acceleration models (A), which account for differences in acceleration and deceleration behaviors, consistently exhibit lower RMSE compared to deceleration models (D). Moreover, the results of the NeMeFIS models is compared with the traditional Linear Regression (LR) (Helly 1959), ANFIS (Kikuchi and Chakroborty 1992) and LSTM (Huang et al. 2018). NeMeFIS is compared with traditional and machine learning benchmarks. Traditional models like Gipps (1981) and GHR (Chandler et al. 1958) lack support for multivariate psychological inputs, while Helly's model (1959) allows them via its linear structure. ANFIS is selected for its technical similarity to NeMeFIS, despite limited cognitive modeling. The number of membership functions for ANFIS were kept the same as the NeMeFIS models. LSTM is included because it also incorporates human memory through temporal dependency modeling and therefore shares conceptual similarity with the NeMeFIS architecture, making it an appropriate benchmark for comparison. The comparison shows that NeMeFIS outperforms the benchmark models in 81 out of 162 cases, while the remaining 81 cases are distributed among the other three model types.

**TABLE 4 Performance (i.e. RMSE) comparison of the NeMeFIS models**

*(a) Training*

*Acceleration*

| Corridor | B | PC | C | M | P | R | AR | T | ALL |
|---|---|---|---|---|---|---|---|---|---|
| AR (HFIS) | 0.11 | 0.10 | 0.11 | 0.12 | 0.14 | 0.28 | 0.10 | - | 0.11 |
| AR (LR) | 0.34 | 0.31 | 0.37 | 0.36 | 0.36 | 0.67 | 0.33 | - | 0.33 |
| AR (LSTM) | 0.16 | 0.14 | 0.16 | 0.12 | 0.19 | 0.26 | 0.14 | - | 0.16 |
| AR (ANFIS) | 0.16 | 0.11 | 0.12 | 0.33 | 0.17 | 0.10 | 0.25 | - | 0.14 |
| CR (HFIS) | 0.08 | 0.10 | 0.12 | 0.08 | 0.08 | 0.11 | 0.12 | - | 0.12 |
| CR (LR) | 0.43 | 0.35 | 0.40 | 0.35 | 0.17 | 0.16 | 0.40 | - | 0.12 |
| CR (LSTM) | 0.10 | 0.11 | 0.08 | 0.15 | 0.07 | 0.17 | 0.14 | - | 0.16 |
| CR (ANFIS) | 0.10 | 0.11 | 0.15 | 0.15 | 0.07 | 0.11 | 0.16 | - | 0.22 |
| RH (HFIS) | 0.11 | 0.22 | 0.15 | 0.15 | 0.10 | 0.23 | 0.12 | 0.13 | 0.09 |
| RH (LR) | 0.64 | 0.13 | 0.64 | 0.33 | 0.31 | 0.15 | 0.13 | 0.32 | 0.40 |
| RH(LSTM) | 0.10 | 0.10 | 0.43 | 0.13 | 0.13 | 0.17 | 0.12 | 0.14 | 0.20 |
| RH (ANFIS) | 0.10 | 0.12 | 0.08 | 0.17 | 0.13 | 0.14 | 0.13 | 0.13 | 0.28 |

*Deceleration*

| Corridor | B | PC | C | M | P | R | AR | T | ALL |
|---|---|---|---|---|---|---|---|---|---|
| AR (HFIS) | 0.12 | 0.02 | 0.24 | 0.10 | 0.04 | 0.22 | 0.09 | - | 0.12 |
| AR (LR) | 0.27 | 0.33 | 0.40 | 0.40 | 0.39 | 0.10 | 0.38 | - | 0.31 |
| AR (LSTM) | 0.19 | 0.14 | 0.22 | 0.14 | 0.10 | 0.29 | 0.17 | - | 0.21 |
| AR (ANFIS) | 0.02 | 0.02 | 0.03 | 0.02 | 0.06 | 0.02 | 0.01 | - | 0.02 |
| CR (HFIS) | 0.01 | 0.05 | 0.01 | 0.08 | 0.11 | 0.02 | 0.03 | - | 0.02 |
| CR (LR) | 0.18 | 0.12 | 0.26 | 0.42 | 0.46 | 0.36 | 0.28 | - | 0.21 |
| CR (LSTM) | 0.24 | 0.20 | 0.27 | 0.23 | 0.19 | 0.22 | 0.23 | - | 0.23 |
| CR (ANFIS) | 0.02 | 0.02 | 0.01 | 0.01 | 0.01 | 0.03 | 0.05 | - | 0.03 |
| RH (HFIS) | 0.02 | 0.11 | 0.19 | 0.11 | 0.09 | 0.13 | 0.10 | 0.10 | 0.12 |
| RH (LR) | 0.47 | 0.62 | 0.41 | 0.43 | 0.25 | 0.47 | 0.42 | 0.37 | 0.41 |
| RH (LSTM) | 0.24 | 0.17 | 0.16 | 0.15 | 0.19 | 0.39 | 0.17 | 0.18 | 0.43 |
| RH (ANFIS) | 0.02 | 0.04 | 0.02 | 0.02 | 0.01 | 0.02 | 0.06 | 0.03 | 0.15 |

*(b) Testing*

*Acceleration*

| Corridor | B | PC | C | M | P | R | AR | T | ALL |
|---|---|---|---|---|---|---|---|---|---|
| AR (HFIS) | 0.10 | 0.11 | 0.08 | 0.10 | 0.15 | 0.30 | 0.09 | - | 0.10 |
| AR (LR) | 0.35 | 0.32 | 0.36 | 0.33 | 0.39 | 0.51 | 0.35 | - | 0.34 |
| AR (LSTM) | 0.29 | 0.28 | 0.35 | 0.29 | 0.39 | 0.40 | 0.23 | - | 0.29 |
| AR (ANFIS) | 0.11 | 0.14 | 0.14 | 0.12 | 0.25 | 0.59 | 0.12 | - | 0.11 |
| CR (HFIS) | 0.06 | 0.09 | 0.16 | 0.07 | 0.29 | 0.11 | 0.13 | - | 0.12 |
| CR (LR) | 0.43 | 0.38 | 0.41 | 0.35 | 0.41 | 0.40 | 0.39 | - | 0.37 |
| CR (LSTM) | 0.29 | 0.18 | 0.35 | 0.45 | 0.25 | 0.15 | 0.21 | - | 0.19 |
| CR (ANFIS) | 0.39 | 0.15 | 0.23 | 0.09 | 0.53 | 0.11 | 0.19 | - | 0.10 |
| RH (HFIS) | 0.12 | 0.17 | 0.13 | 0.11 | 0.13 | 0.30 | 0.23 | 0.13 | 0.10 |
| RH (LR) | 0.69 | 0.17 | 0.66 | 0.36 | 0.27 | 0.32 | 0.68 | 0.17 | 0.19 |
| RH (LSTM) | 0.36 | 0.34 | 0.48 | 0.49 | 0.35 | 0.40 | 0.25 | 0.31 | 0.37 |
| RH (ANFIS) | 0.39 | 0.13 | 0.21 | 0.20 | 0.12 | 0.42 | 0.13 | 0.14 | 0.87 |

*Deceleration*

| Corridor | B | PC | C | M | P | R | AR | T | ALL |
|---|---|---|---|---|---|---|---|---|---|
| AR (HFIS) | 0.11 | 0.01 | 0.06 | 0.08 | 0.03 | 0.25 | 0.08 | - | 0.12 |
| AR (LR) | 0.32 | 0.33 | 0.32 | 0.42 | 0.41 | 0.54 | 0.34 | - | 0.36 |
| AR (LSTM) | 0.20 | 0.22 | 0.50 | 0.34 | 0.31 | 0.61 | 0.21 | - | 0.28 |
| AR (ANFIS) | 0.03 | 0.02 | 0.04 | 0.02 | 0.06 | 0.02 | 0.02 | - | 0.02 |
| CR (HFIS) | 0.07 | 0.01 | 0.04 | 0.06 | 0.08 | 0.02 | 0.07 | - | 0.08 |
| CR (LR) | 0.25 | 0.31 | 0.43 | 0.42 | 0.47 | 0.39 | 0.45 | - | 0.29 |
| CR (LSTM) | 0.27 | 0.18 | 0.67 | 0.50 | 0.45 | 0.26 | 0.40 | - | 0.26 |
| CR (ANFIS) | 0.02 | 0.01 | 0.01 | 0.01 | 0.01 | 0.03 | 0.04 | - | 0.03 |
| RH (HFIS) | 0.02 | 0.04 | 0.32 | 0.10 | 0.09 | 0.04 | 0.03 | 0.08 | 0.11 |
| RH (LR) | 0.55 | 0.59 | 0.37 | 0.37 | 0.37 | 0.53 | 0.42 | 0.38 | 0.44 |
| RH (LSTM) | 0.36 | 0.26 | 0.29 | 0.26 | 0.45 | 0.64 | 0.29 | 0.38 | 0.35 |
| RH (ANFIS) | 0.02 | 0.04 | 0.23 | 0.03 | 0.58 | 0.11 | 0.04 | 0.02 | 0.22 |

*(c) Validation*

*Acceleration*

| Corridor | B | PC | C | M | P | R | AR | T | ALL |
|---|---|---|---|---|---|---|---|---|---|
| AR (HFIS) | 0.11 | 0.10 | 0.22 | 0.10 | 0.25 | 0.51 | 0.11 | - | 0.11 |

*Deceleration*

| Corridor | B | PC | C | M | P | R | AR | T | ALL |
|---|---|---|---|---|---|---|---|---|---|
| AR (HFIS) | 0.11 | 0.01 | 0.20 | 0.02 | 0.13 | 0.44 | 0.10 | - | 0.13 |

| *Acceleration* | | | | | | | | | | *Deceleration* | | | | | | | | | |
|---|---|---|---|---|---|---|---|---|---|---|---|---|---|---|---|---|---|---|---|
| **Corridor** | **B** | **PC** | **C** | **M** | **P** | **R** | **AR** | **T** | **ALL** | **Corridor** | **B** | **PC** | **C** | **M** | **P** | **R** | **AR** | **T** | **ALL** |
| **AR (LR)** | 0.36 | 0.32 | 0.37 | 0.36 | 0.39 | 0.57 | 0.31 | - | 0.33 | **AR (LR)** | 0.22 | 0.37 | 0.30 | 0.42 | 0.42 | 0.54 | 0.38 | - | 0.39 |
| **AR (LSTM)** | 0.32 | 0.28 | 0.35 | 0.36 | 0.62 | 0.29 | 0.22 | - | 0.28 | **AR (LSTM)** | 0.31 | 0.28 | 0.37 | 0.39 | 0.51 | 0.62 | 0.20 | - | 0.24 |
| **AR (ANFIS)** | 0.43 | 0.11 | 0.29 | 0.22 | 0.09 | 0.46 | 0.31 | - | 0.12 | **AR (ANFIS)** | 0.02 | 0.02 | 0.07 | 0.02 | 0.04 | 0.02 | 0.01 | - | 0.02 |
| **CR (HFIS)** | **0.09** | **0.09** | 0.13 | **0.05** | 0.13 | **0.12** | 0.12 | - | **0.11** | **CR (HFIS)** | 0.09 | **0.02** | 0.08 | **0.01** | 0.13 | **0.02** | **0.03** | - | 0.08 |
| **CR (LR)** | 0.42 | 0.40 | 0.38 | 0.37 | 0.22 | 0.41 | 0.39 | - | 0.25 | **CR (LR)** | 0.31 | 0.45 | 0.40 | 0.42 | 0.47 | 0.44 | 0.40 | - | 0.36 |
| **CR (LSTM)** | 0.20 | 0.22 | 0.35 | 0.41 | 0.26 | 0.26 | 0.59 | - | 0.16 | **CR (LSTM)** | 0.37 | 0.30 | 0.51 | 0.36 | 0.42 | 0.30 | 0.31 | - | 0.20 |
| **CR (ANFIS)** | 0.09 | 0.15 | 0.12 | 0.14 | 0.10 | 0.78 | 0.09 | - | 0.19 | **CR (ANFIS)** | 0.01 | 0.02 | 0.01 | 0.01 | 0.05 | 0.02 | 0.04 | - | 0.03 |
| **RH (HFIS)** | 0.26 | **0.20** | 0.41 | **0.10** | **0.10** | 0.44 | 0.44 | 0.19 | **0.10** | **RH (HFIS)** | **0.08** | 0.10 | 0.63 | 0.18 | **0.10** | 0.18 | 0.10 | 0.14 | **0.12** |
| **RH (LR)** | 0.68 | 0.18 | 0.49 | 0.31 | 0.26 | 0.17 | 0.62 | 0.36 | 0.38 | **RH (LR)** | 0.56 | 0.66 | 0.22 | 0.49 | 0.35 | 0.46 | 0.42 | 0.38 | 0.32 |
| **RH (LSTM)** | 0.35 | 0.39 | 0.30 | 0.39 | 0.35 | 0.32 | 0.32 | 0.24 | 0.43 | **RH (LSTM)** | 0.42 | 0.36 | 0.56 | 0.41 | 0.32 | 0.45 | 0.33 | 0.39 | 0.42 |
| **RH (ANFIS)** | 0.21 | 0.59 | 0.25 | 0.88 | 0.16 | 0.12 | 0.13 | 0.12 | 0.42 | **RH(ANFIS)** | 0.08 | 0.03 | 0.13 | 0.07 | 0.49 | 0.12 | 0.06 | 0.03 | 0.22 |

*Vehicle Types: B = Bus, PC =Passenger Car, C = Covered van, M = Motorcycle, P = Pick-up, R = Rickshaw, AR = Auto-Rickshaw, T= Truck; HFIS = Modeled with NeMeFIS; *hyphen (-) indicates absence of sample in the particular vehicle type*

NeMeFIS (HFIS) performs best mainly in deceleration scenarios across all corridors (AR, CR, RH) and vehicle types, showing consistently lower RMSE in training, testing, and validation. Its advantage is strongest for passenger cars, buses, motorcycles, and mixed traffic, particularly on collector roads and rural highways, where braking involves higher cognitive complexity. HFIS also shows strong performance in acceleration on CR and RH, though with smaller margins. This superiority stems from its asymmetric, memory-informed fuzzy structure, which captures nonlinear and context-dependent human driving decisions more effectively than LR, ANFIS, and LSTM.

**Table 5** summarizes the number of fuzzy inference system (FIS) rules in the Neuro-Memory Fuzzy Inference System (NeMeFIS) across different driving scenarios. The results reveal that acceleration decisions involve fewer rules under procedural memory, emphasizing skill-based responses, while declarative memory consistently yields the highest rule counts, reflecting its role in complex cognitive reasoning. Deceleration involves significantly higher rule complexity, especially under declarative and combined memory types.

**TABLE 5 Number of Rules in the NeMeFIS model in different scenarios and different vehicle types**

| **CT** | **FIS** | *Acceleration (**A**)* | | | | | | | | | *Deceleration (**D**)* | | | | | | | | |
|---|---|---|---|---|---|---|---|---|---|---|---|---|---|---|---|---|---|---|---|
| | | **B** | **PC** | **C** | **M** | **P** | **R** | **AR** | **T** | **A** | **B** | **PC** | **C** | **M** | **P** | **R** | **AR** | **T** | **A** |
| **AR** | **P** | 7 | 7 | 8 | 8 | 8 | 10 | 9 | 0 | 17 | 11 | 8 | 9 | 11 | 8 | 10 | 9 | 0 | 26 |
| | **W** | 8 | 11 | 10 | 7 | 8 | 9 | 7 | 0 | 21 | 9 | 9 | 6 | 9 | 10 | 10 | 10 | 0 | 24 |
| | **E** | 9 | 9 | 7 | 9 | 9 | 9 | 9 | 0 | 21 | 8 | 9 | 7 | 10 | 11 | 10 | 10 | 0 | 25 |
| | **S** | 27 | 32 | 42 | 43 | 36 | 42 | 34 | 0 | 115 | 37 | 37 | 38 | 40 | 39 | 40 | 41 | 0 | 133 |
| | **D** | 160 | 151 | 167 | 167 | 169 | 195 | 170 | 0 | 332 | 190 | 170 | 187 | 199 | 190 | 177 | 192 | 0 | 448 |
| | **C** | 71 | 65 | 69 | 74 | 72 | 94 | 86 | 0 | 128 | 83 | 84 | 90 | 93 | 97 | 76 | 95 | 0 | 166 |
| **CR** | **P** | 9 | 9 | 11 | 11 | 10 | 9 | 10 | 0 | 14 | 9 | 8 | 8 | 10 | 10 | 11 | 10 | 0 | 23 |
| | **W** | 9 | 9 | 8 | 8 | 6 | 10 | 7 | 0 | 17 | 12 | 7 | 9 | 10 | 10 | 11 | 9 | 0 | 25 |
| | **E** | 8 | 10 | 8 | 11 | 8 | 10 | 9 | 0 | 18 | 9 | 8 | 10 | 12 | 6 | 9 | 10 | 0 | 21 |
| | **S** | 33 | 44 | 35 | 41 | 33 | 36 | 37 | 0 | 110 | 38 | 30 | 37 | 37 | 41 | 39 | 42 | 0 | 129 |
| | **D** | 155 | 196 | 155 | 185 | 171 | 171 | 163 | 0 | 300 | 223 | 135 | 200 | 210 | 218 | 171 | 211 | 0 | 427 |
| | **C** | 68 | 84 | 67 | 86 | 83 | 79 | 84 | 0 | 116 | 99 | 59 | 90 | 100 | 98 | 78 | 103 | 0 | 149 |
| **RH** | **P** | 9 | 7 | 9 | 7 | 10 | 8 | 10 | 10 | 19 | 10 | 11 | 7 | 8 | 8 | 9 | 9 | 9 | 18 |
| | **W** | 10 | 11 | 11 | 9 | 10 | 10 | 8 | 7 | 18 | 10 | 9 | 8 | 8 | 7 | 9 | 11 | 11 | 18 |
| | **E** | 10 | 8 | 9 | 8 | 11 | 8 | 9 | 8 | 18 | 9 | 7 | 10 | 10 | 9 | 8 | 10 | 8 | 22 |
| | **S** | 41 | 38 | 35 | 41 | 42 | 34 | 42 | 36 | 116 | 36 | 36 | 37 | 32 | 40 | 39 | 36 | 42 | 104 |
| | **D** | 184 | 185 | 156 | 183 | 209 | 200 | 192 | 185 | 315 | 187 | 199 | 176 | 162 | 149 | 216 | 212 | 194 | 332 |
| | **C** | 86 | 79 | 84 | 86 | 107 | 89 | 83 | 90 | 111 | 92 | 97 | 89 | 74 | 66 | 109 | 98 | 83 | 129 |

** CT = Corridor Type, FIS = Fuzzy inference System, FIS types: P = Procedural Memory, W = Working Memory, E = Episodic memory, S = Semantic memory, D = Declarative memory, C = Combined Memory;*

**Figure 2** illustrates the sensitivity of the Neuro Memory Fuzzy Inference System (NeMeFIS) model's acceleration magnitude (normalized) to various normalized input variables across different traffic scenarios. S-L Spacing strongly influences acceleration magnitude at lower normalized values (0–0.3), emphasizing the need for safe following distances in proximity-dominated conditions. Risk Perception consistently remains a dominant variable across all traffic scenarios, reinforcing its critical role in cautious and human-like decision-making. In AR and RH settings, Speed-S and Relative Speed S-NL exhibit noticeable sensitivity, emphasizing their importance in managing speed mismatches and overtaking behavior.

**Figure 3** illustrates the variable importance across vehicle types and road corridors during acceleration and deceleration. This figure reveals that risk perception dominates during acceleration, especially on Arterial (AR) and Collector Roads (CR), with covered vans and motorcycles showing the highest sensitivity due to their contrasting vulnerabilities. In Rural Highways (RH), however, angular spacing becomes more critical for buses aiming to overtake. During deceleration, the emphasis shifts from risk to linear and angular spacing, reflecting adaptive strategies across corridors. In AR, lead spacing ranks highest, followed by angular cues for potential lane changes. In CR, angular distance to the lead vehicle dominates due to weak lane discipline. Finally, in RH corridors, relative spacing with the lead vehicle is most influential, reflecting disciplined traffic flow and fewer overtaking options.

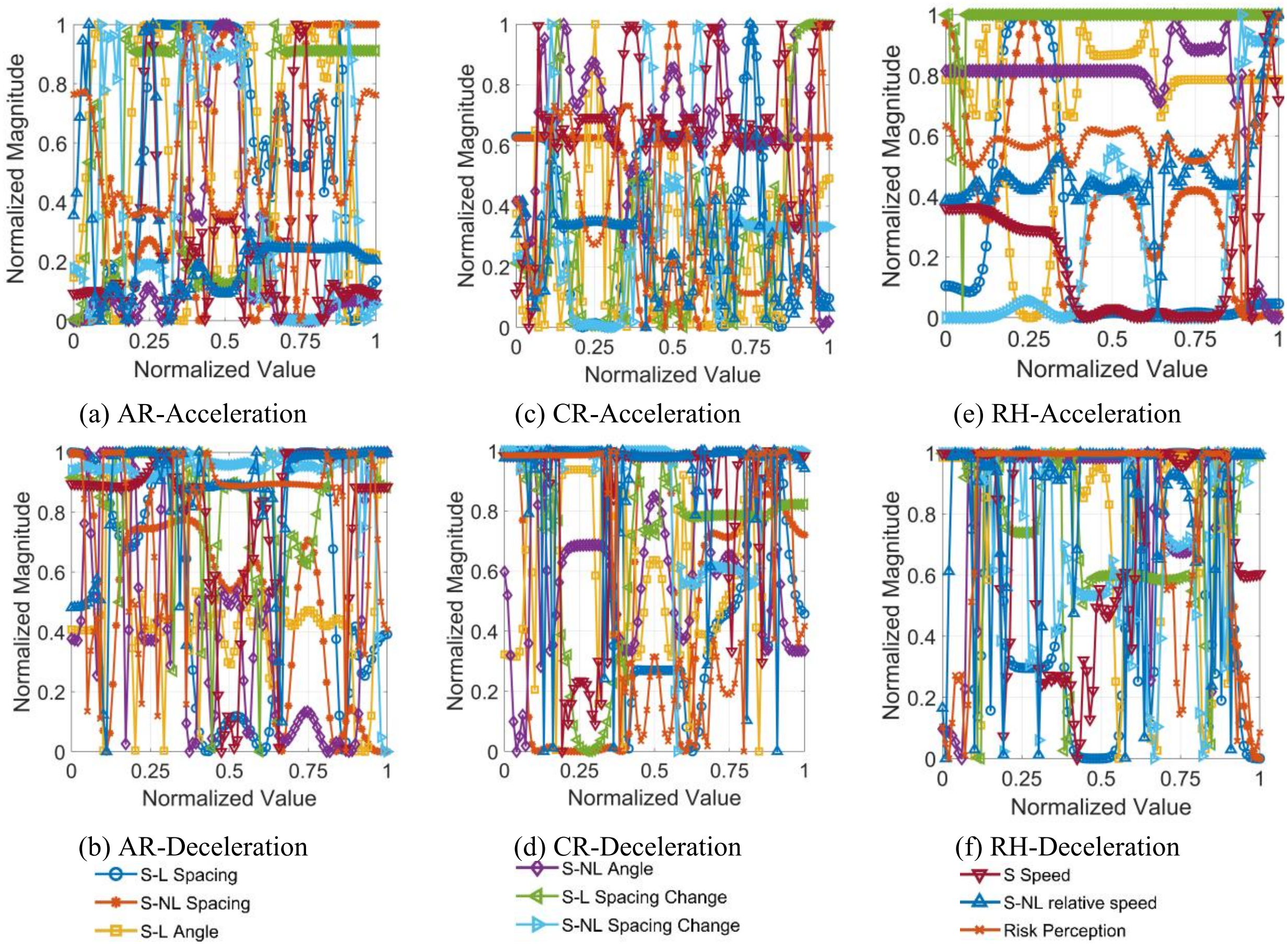

(a) AR-Acceleration (c) CR-Acceleration (e) RH-Acceleration

(b) AR-Deceleration (d) CR-Deceleration (f) RH-Deceleration

**Figure 2 Sensitivity of Different Variables in NeMeFIS Models**

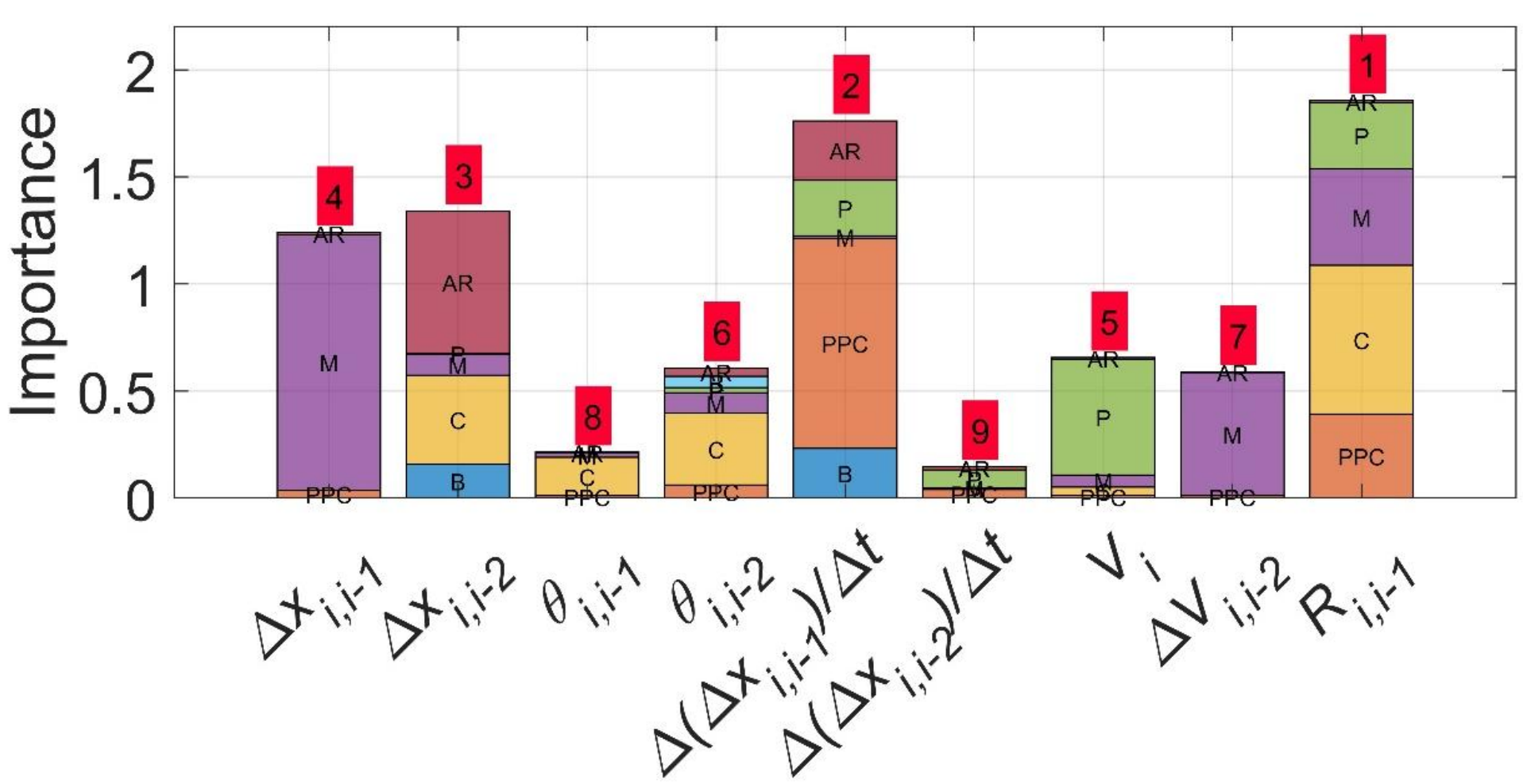

(a) AR-Acceleration

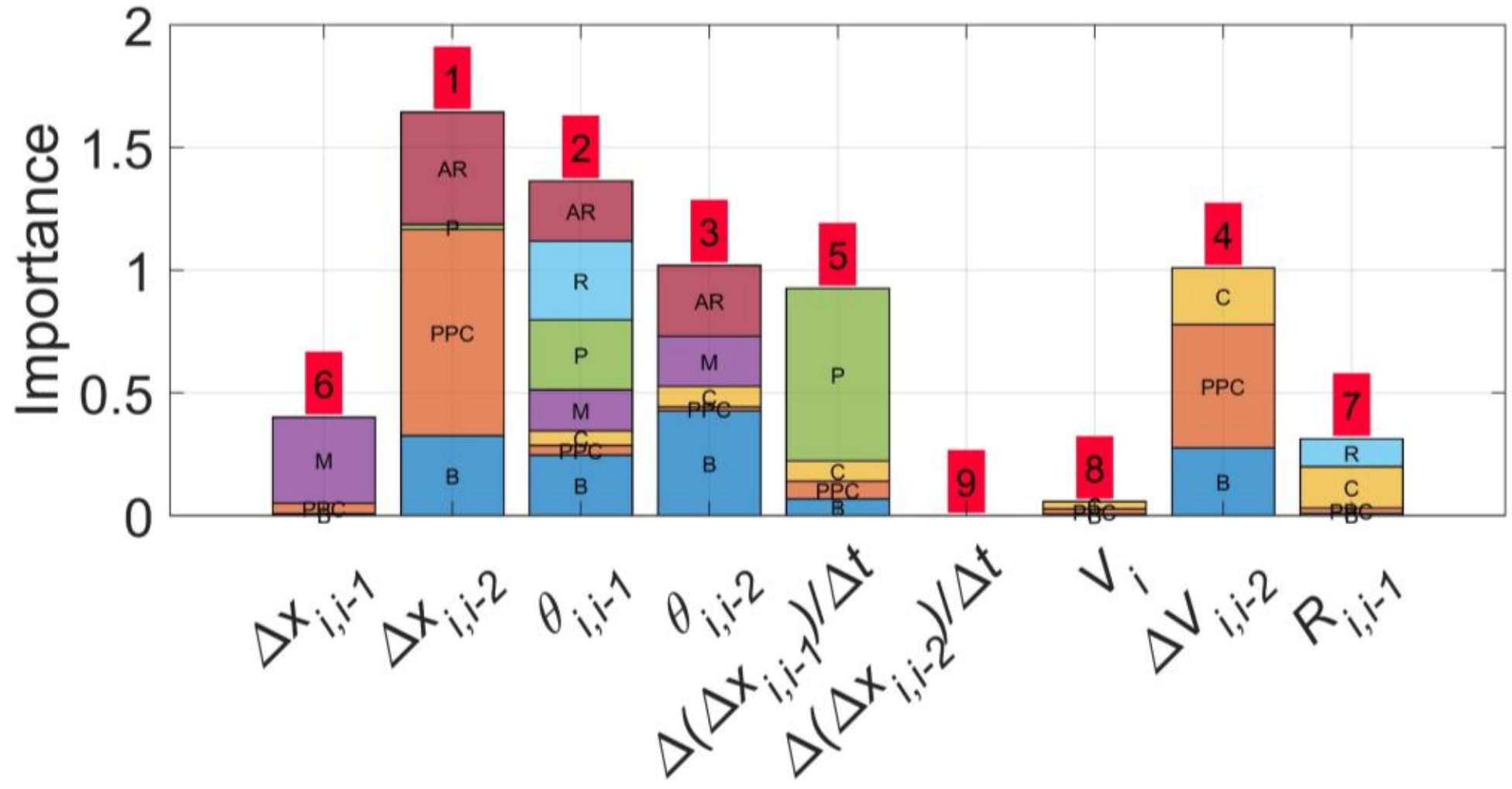

(b) AR-Deceleration

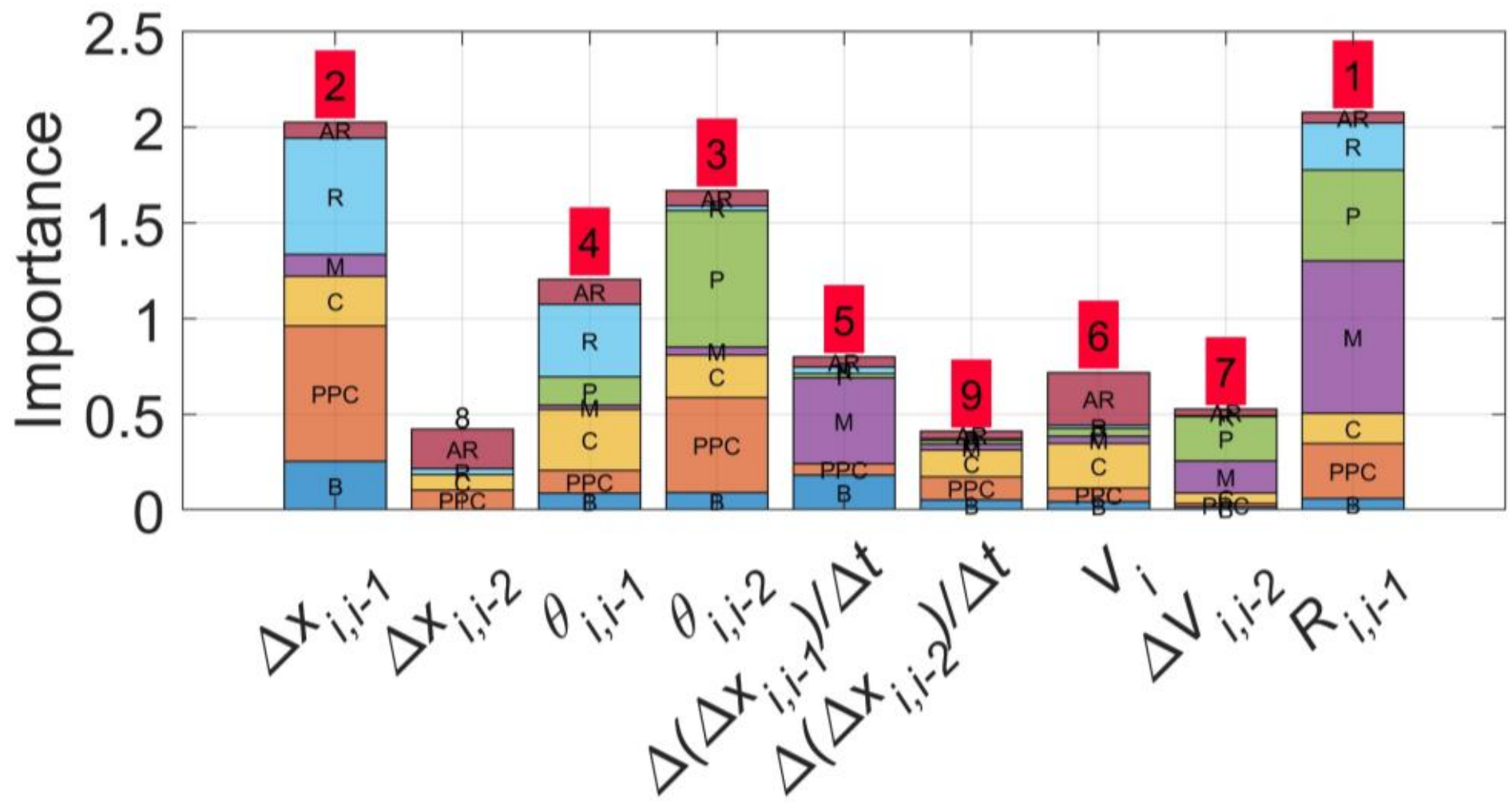

(c) CR-Acceleration

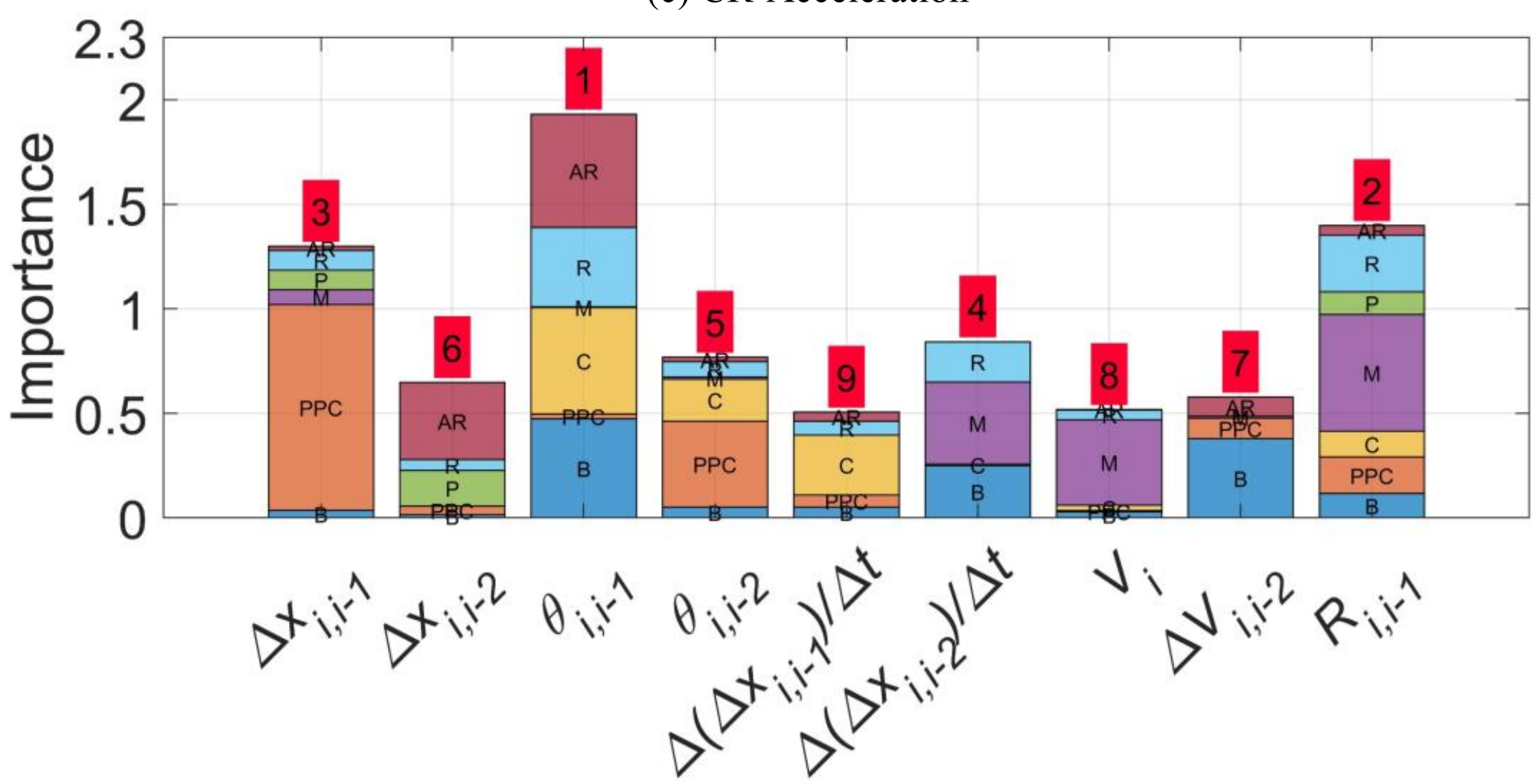

(d) CR-Deceleration

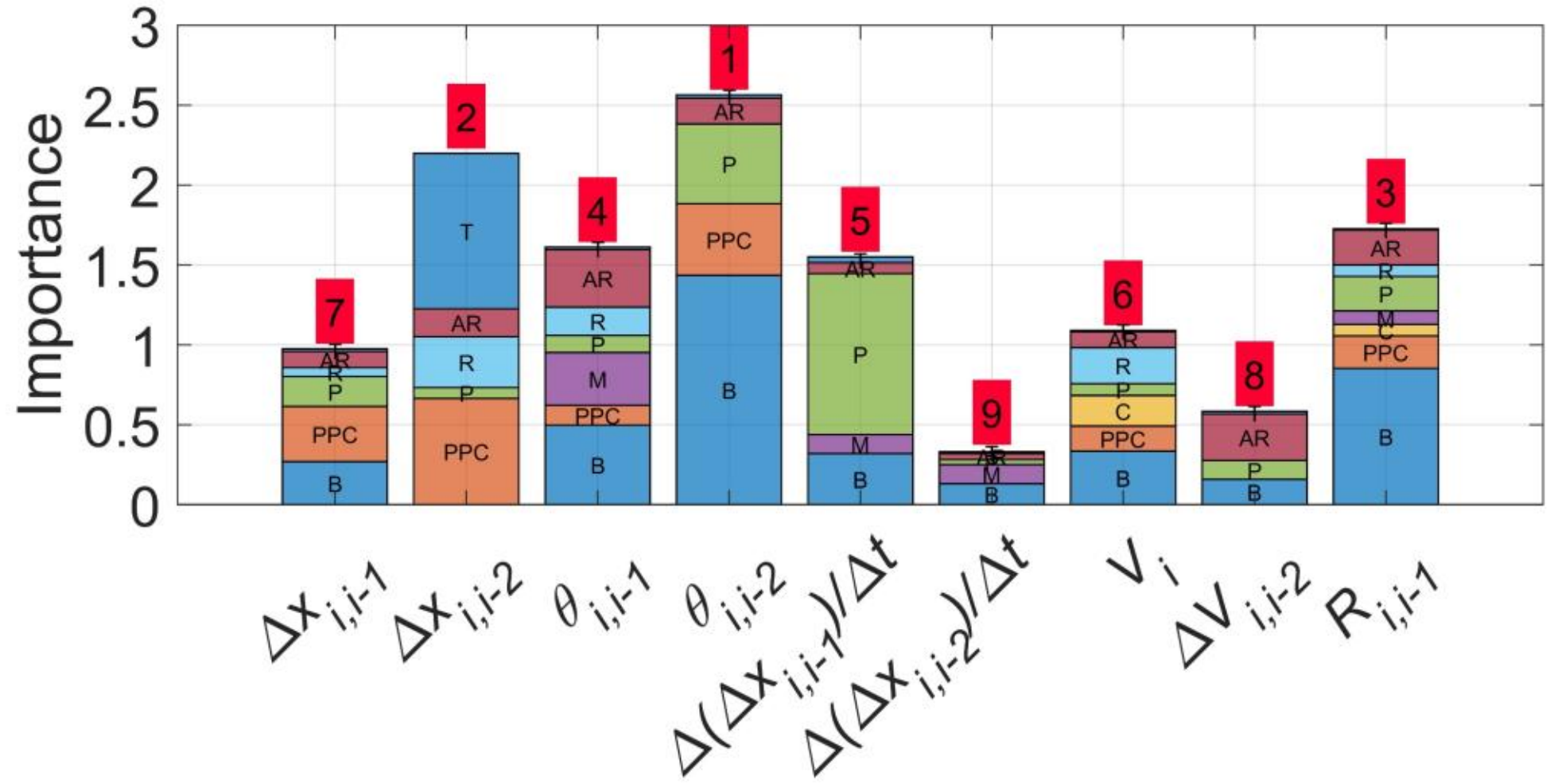

(e) RH-Acceleration

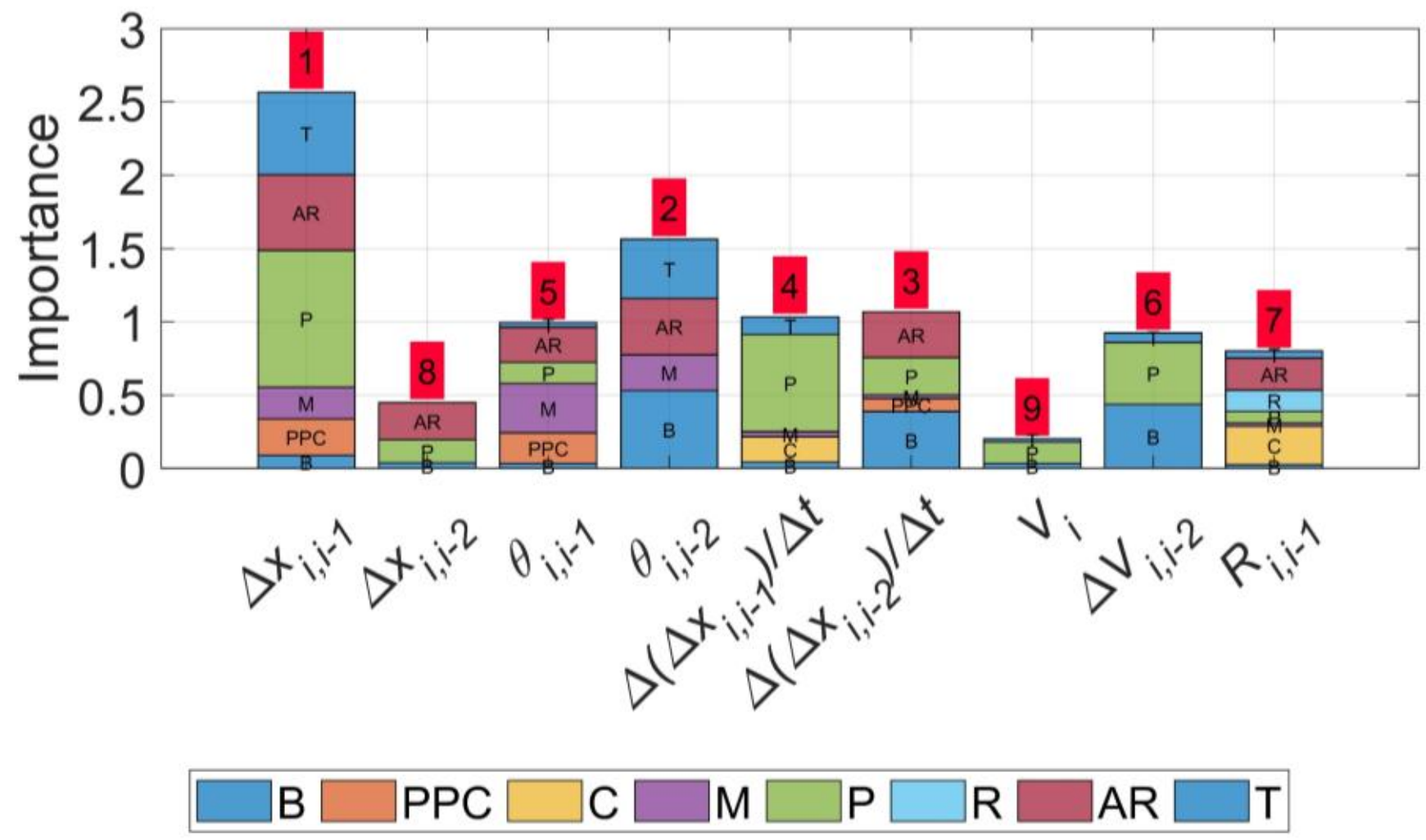

(f) RH-Deceleration

**Figure 3 Variable Importance for Different Scenarios and Vehicles**

**Figure 4** illustrates the probabilistic influence of various memory types. The centroid values and polygon areas represent the intensity and spread of each memory's contribution. This figure shows that Procedural (P) memory dominates acceleration on Arterial and Collector Roads (AR, CR), with high centroid values (e.g., P3 on AR) indicating skill-based control at higher speeds. Working (W) memory is also crucial during AR and CR acceleration, supporting real-time adaptations (e.g., clusters W2, W3). During AR deceleration, Semantic (S) memory leads (e.g., S1), aiding in signal anticipation through generalized knowledge. On Collector Roads, Declarative (D) memory dominates deceleration (e.g., D3), reflecting the need to synthesize spatial and experiential inputs in chaotic conditions. In Rural Highways (RH), Semantic memory governs acceleration (e.g., S2, S3), while Declarative memory guides deceleration (e.g., D1, D3) to handle unpredictable hazards. In summary, Procedural memory drives acceleration, while Declarative and Semantic memories dominate deceleration, shaped by road type and traffic complexity.

Furthermore, the stability of the model was evaluated using numerical simulation (results shown in **Figure 5**). The simulation environment was configured as follows: total simulation time of 100 seconds, a time step of 0.1 seconds, a distance step of 0.1 meters, one lane with a width of 3.65 meters, a road length of 1000 meters, a flow rate resolution of 10 seconds, a flow rate of 1500 vehicles per hour, vehicle composition based on the corridor type, and a perception-reaction time of 1.5 seconds. Simulations were conducted across all three corridor types to assess the stability of the model. Simulations across three corridor types confirmed the model's stability under controlled acceleration, deceleration, and shockwave-inducing conditions. Vehicle trajectories showed no FIFO violations or abrupt speed changes, with minor oscillations damped over time, indicating stable behavior. Emergency braking spikes observed in **Figure 5** stemmed from boundary conditions during vehicle entry, not model deficiencies. Thus, the results affirm the proposed model's robustness and resistance to instability in diverse traffic scenarios.

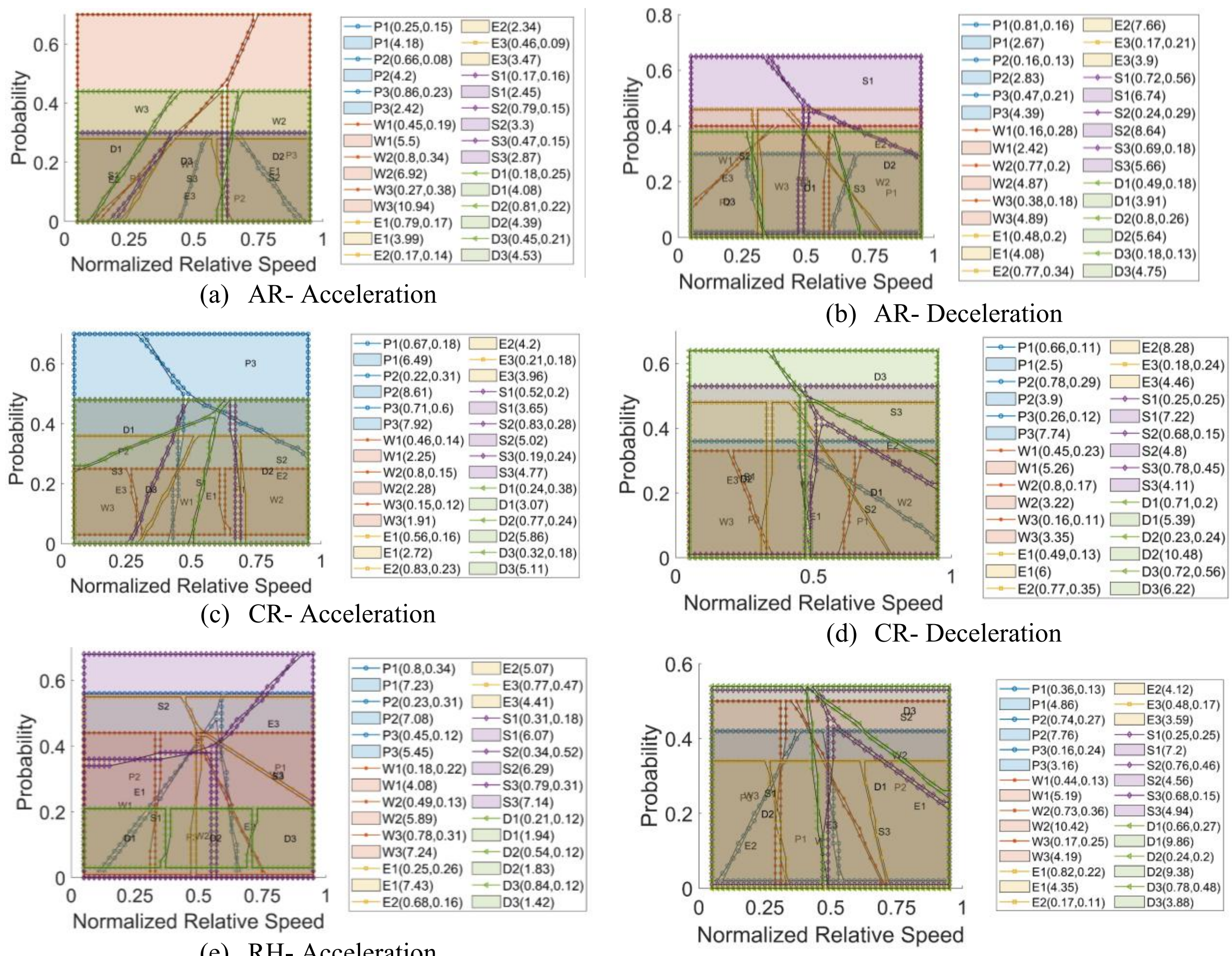

(a) AR- Acceleration

(b) AR- Deceleration

(c) CR- Acceleration

(d) CR- Deceleration

(e) RH- Acceleration

(f) RH- Deceleration

**Figure 4** Contribution of Memory Types in Decision Making at Different Relative Speed Level to Choose Magnitude.

The transferability of the proposed Neuro Memory Fuzzy Inference System (NeMeFIS) architecture was trained and evaluated using sampled trajectory datasets from (a) freeway US-101, sourced from the Next Generation Simulation (US Department of Transportation 2007) program (**Figure 6(a)**) and, (b) Argoverse 2 (Wilson et al. 2023) which facilitates CAV trajectories (**Figure 6(b)**). The NeMeFIS architecture was found to be transferable with considerable accuracy and sensitivity in both the datasets.

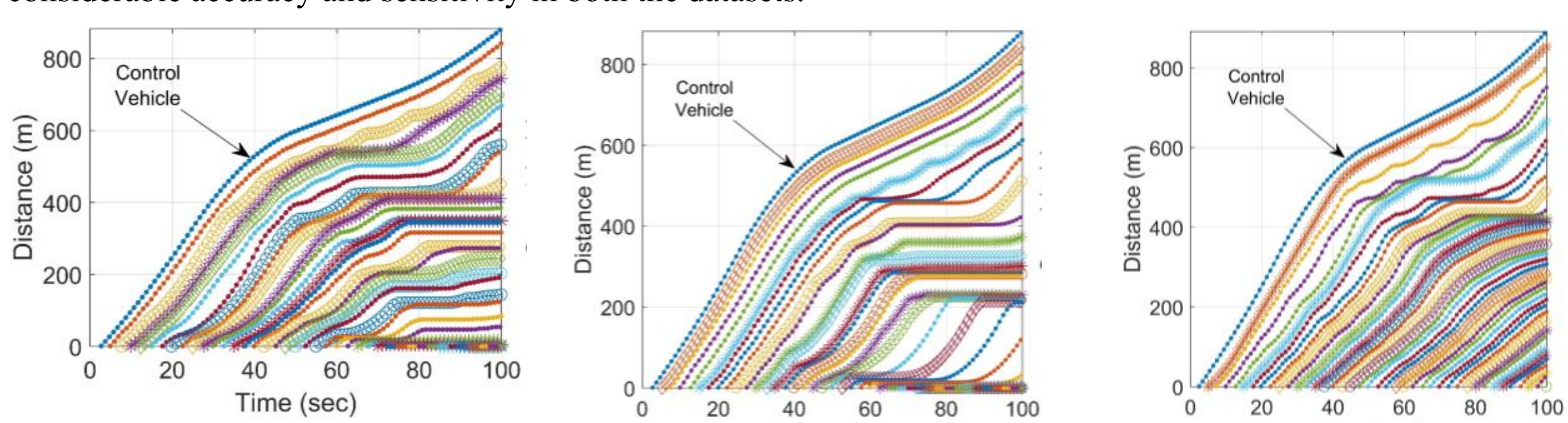

(a)

(c)

(e)

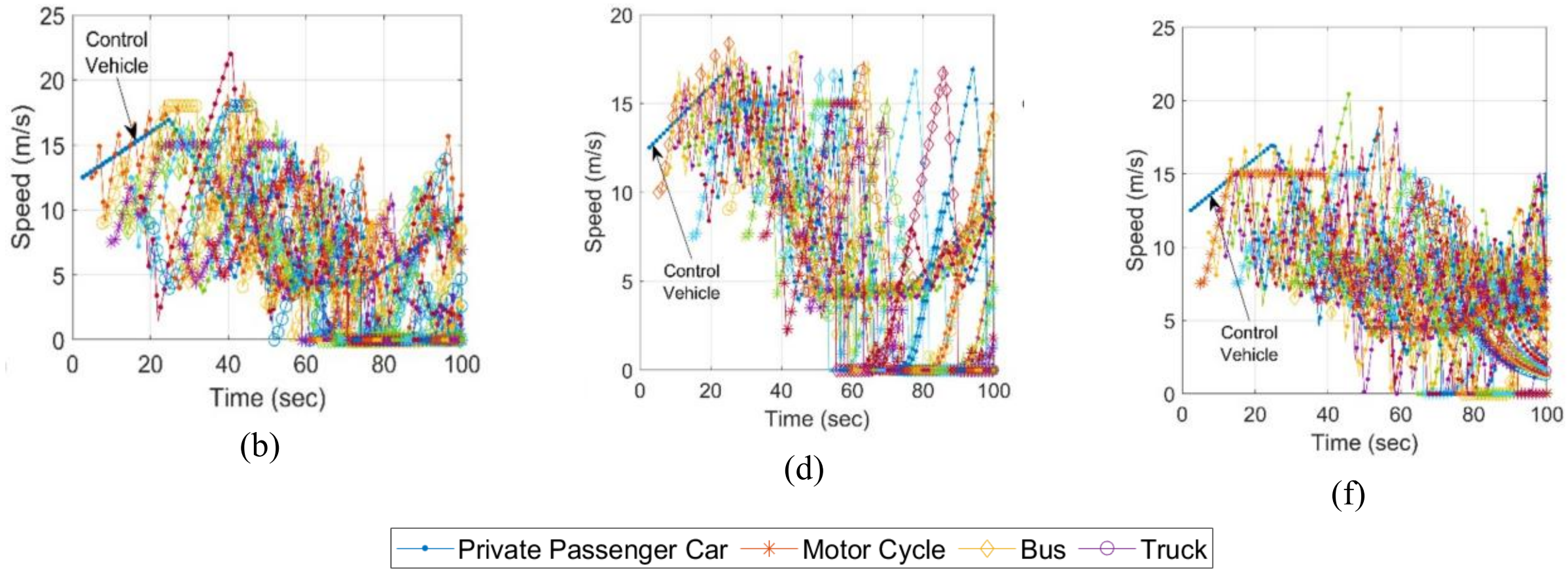

**Figure 5 Numerical Simulation of the NeMeFIS Models for different vehicle types (a)-(b) AR; (c)-(d) CR; (e)-(f) RH**

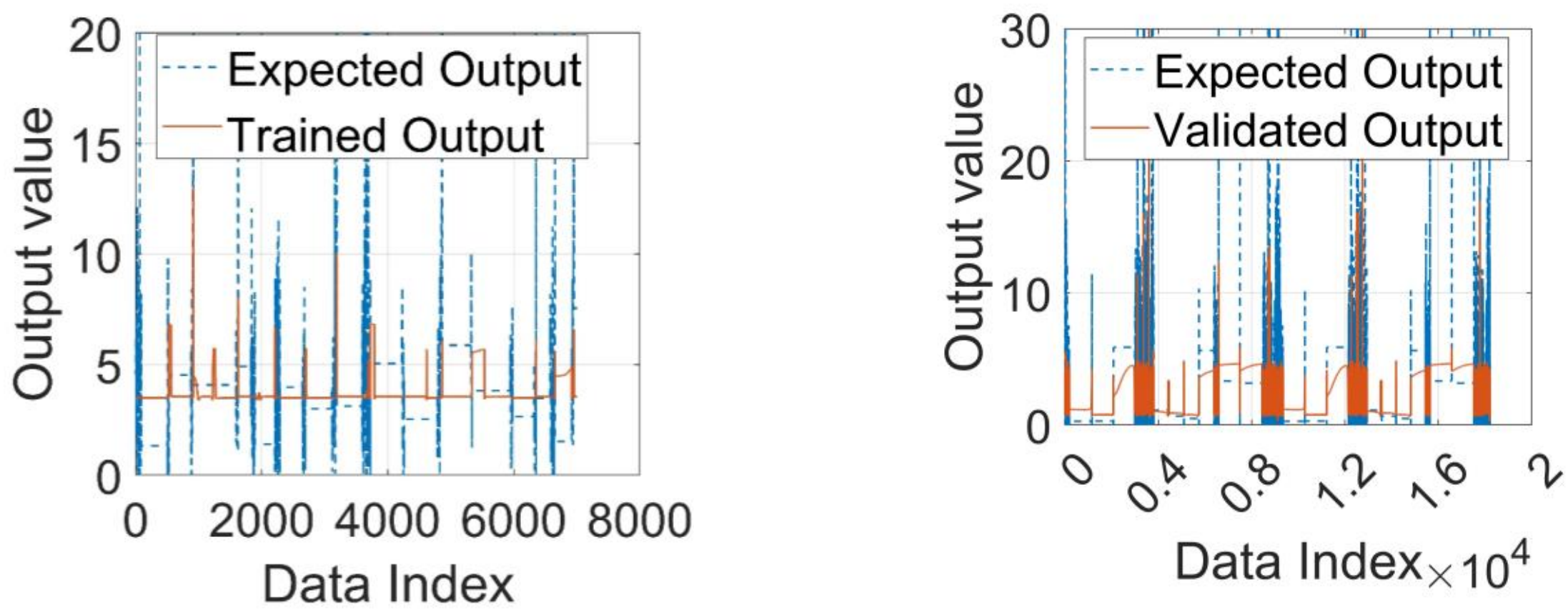

(a) (b)

**Figure 6 Transferability and Convergence of NeMeFIS Architecture in Different Datasets (a) NGSIM (b) Argoverse 2 (CAV)**

## CONCLUSIONS

This study presents the Neuro Memory Fuzzy Inference System (NeMeFIS), a novel model that simulates human-like vehicle-following behavior by incorporating five types of memory: procedural, working, episodic, semantic, and declarative. Acceleration is primarily influenced by procedural and working memory, enabling skill-based and adaptive responses, while deceleration relies on declarative and semantic memory, reflecting complex cognitive demands. The model reveals higher cognitive load for rural highways and larger vehicles, as indicated by increased fuzzy rule counts. Applied on UAV-based (heterogeneous) traffic data and evaluated on US101 (homogenous) and Argoverse 2 (CAV), NeMeFIS demonstrates robustness and transferability. Notably, NeMeFIS consistently outperforms linear regression, ANFIS, and LSTM models, particularly in asymmetric deceleration regimes, highlighting its practical advantage over existing approaches. Future improvements should incorporate extreme case data, tailored loss functions, or models with increased membership functions as well as rules. Expanding the framework to include psychological and physiological factors like stress and fatigue will enhance its realism. NeMeFIS holds potential for driver behavior modification programs and integration into CAVs for context-aware, human-like, and safety-oriented decision-making.

## ACKNOWLEDGMENTS

The author(s) acknowledge the use of Grammarly (Grammarly Inc.), an AI-based writing assistance tool, during the preparation of this manuscript. Grammarly was used solely for grammar, spelling, punctuation, and clarity checks to improve the readability of the written text. The tool was not used to generate original content, ideas, data, analysis, or conclusions presented in this paper. All intellectual content, research design, data analysis, and interpretations remain the sole responsibility of the author(s).

## AUTHOR CONTRIBUTIONS

The authors confirm contribution to the paper as follows: study conception and design: N. Haque, M. Hadiuzzaman; data collection: N. Haque; deep learning architecture development: N. Haque, analysis and interpretation of results: N. Haque; draft manuscript preparation: N. Haque, M. A. Raihan. All authors reviewed the results and approved the final version of the manuscript.

## DECLARATION OF CONFLICTING INTERESTS

The authors declared no potential conflicts of interest with respect to the research, authorship, and/or publication of this article.

## FUNDING

This research work is supported by the Committee for Advanced Studies and Research (CASR) (Grant No. 371(16)) of Bangladesh University of Engineering and Technology (BUET).